\pdfoutput=1

\documentclass[11pt]{article}

\usepackage[preprint]{acl}

\usepackage{times}
\usepackage{latexsym}

\usepackage[T1]{fontenc}

\usepackage[utf8]{inputenc}

\usepackage{microtype}

\usepackage{inconsolata}

\usepackage{graphicx}

\usepackage{titlesec}
\titlespacing{\paragraph}{0pt}{0.3\baselineskip}{0.1\baselineskip}

\title{CALE : Concept-Aligned Embeddings for Both Within-Lemma and Inter-Lemma Sense Differentiation}

\author{
 \textbf{Bastien Li\'etard\textsuperscript{1}} \and
 \textbf{Gabriel Loiseau\textsuperscript{1,2}},
\\
 \textsuperscript{1}University of Lille, Inria, CNRS, Centrale Lille, UMR 9189 - CRIStAL, F-59000 Lille, France \\
 \textsuperscript{2}Hornetsecurity, Hem, France
\\
 \texttt{first\_name.last\_name@inria.fr}
\\\small \textit{Authors contributed equally}.
}

\usepackage[dvipsnames]{xcolor}
\usepackage{booktabs, colortbl}
\usepackage{lipsum}
\usepackage{amsthm, amsmath, amssymb}
\usepackage{adjustbox}

\newcommand{\occany}[1][j]{o_{#1}}
\newcommand{\occ}[2][w]{o_{#1,#2}}

\newcommand{\embany}[1][j]{e_{#1}}

\newcommand{\embT}[3][w]{e_{#1,#2}^{#3}}

\begin{document}
\maketitle
\begin{abstract}
Lexical semantics is concerned with both the multiple senses a word can adopt in different contexts, and the semantic relations that exist between meanings of different words. To investigate them, Contextualized Language Models are a valuable tool that provides context-sensitive representations that can be used to investigate lexical meaning. Recent works like XL-LEXEME have leveraged the task of Word-in-Context to fine-tune them to get more semantically accurate representations, but Word-in-Context only compares occurrences of the same lemma, limiting the range of captured information. In this paper, we propose an extension, Concept Differentiation, to include inter-words scenarios. We provide a dataset for this task, derived from SemCor data. Then we fine-tune several representation models on this dataset. We call these models  Concept-Aligned Embeddings (CALE). By challenging our models and other models on various lexical semantic tasks, we demonstrate that the proposed models provide efficient multi-purpose representations of lexical meaning that reach best performances in our experiments. We also show that CALE's fine-tuning brings valuable changes to the spatial organization of embeddings.
\end{abstract}

\section{Introduction}

Research in computational lexical semantics has relied on contextualized embeddings to study word meaning in context \citep{neidlein-etal-2020-analysis, chronis-erk-2020-bishop, apidianaki-gari-soler-2021-dolphins, yu-xu-2023-word, li-etal-2024-investigating}, with applications extending beyond traditional tasks, such as in debate modeling \citep{gari-soler-etal-2023-measuring} and political discourse analysis \citep{boholm-etal-2024-political}. However, most approaches use pre-trained models like BERT \citep{devlin-etal-2019-bert} or XLM-R \citep{conneau-etal-2020-unsupervised} without fine-tuning specifically for lexical semantics, due to limited annotated data.

While several models are fine-tuned for the Word-in-Context (WiC) task \citep{pilehvar-camacho-collados-2019-wic, liu-etal-2021-mirrorwic, cassotti-etal-2023-xl, mosolova-etal-2024-injecting}, WiC in its traditional definition only captures within-word meaning variation (i.e. the word's senses) and ignores inter-word semantic relations. Consequently, models trained on WiC may struggle to capture the broader structure of lexical meaning.

To reach a higher perspective, the focus must shift from word senses to \textit{semantic concepts}. Throughout this paper, we distinguish senses and concepts as follows: a word sense is a conventionalized way to use a particular lemma in relevant contexts, it characterizes the use of a specific word for an intended meaning; concepts are mental representations of categories of objects, events, acts and ideas, and we use words to refer to them \citep{murphy-2002-concepts}. In context, the meaning of a word is the concept it refers to, and a word sense is a pointer to a concept. Our concepts are equivalent to Wikipedia entries, or WordNet synsets \citep{miller-1995-wordnet} as made in \citet{lietard-etal-2024-word}.

In this paper, we are interested in the use of contextualized models to obtain multi-purpose (as opposed to ``task-specific'') vector representation of word meaning. Word occurrences that instantiate the same concept in respective contexts should have similar representations, while occurrences of words that refer to different concepts should have dissimilar embeddings. A model with this desired property would be useful in studies of the mapping between word forms and concepts, like \citet{haspelmath-2023-coexpression} for instance, or in any computational lexical semantic applications mentioned above.

We focus on \textit{synonymy} (different words having the same meaning, instantiating the same concept) and \textit{polysemy} (a single word that can refer to multiple concepts) and we leave to future work the integration of other relations such as hyponymy.

First, we introduce Concept Differentiation, a task that generalizes WiC to compare contextualized meanings of both same-lemma and cross-lemma word pairs. Given two occurrences, the task is to determine whether they instantiate the same semantic concept. Like WiC is to Word Sense Induction, Concept Differentiation is a binary classification form of Concept or Frame Induction \citep{lietard-etal-2024-word, qasemizadeh-etal-2019-semeval}.

To support this task, we construct a new dataset: SPCD (\underline{S}emcor \underline{P}airs for \underline{C}oncept \underline{D}ifferentiation), drawn from SemCor \citep{miller-1995-wordnet}, including both intra- and inter-lemma pairs. This enables evaluation of lexical meaning representations at both local (sense-level) and global (concept-level) scales.

Then, we develop CALE (\underline{C}oncept-\underline{Al}igned \underline{E}mbeddings), a family of token-level models fine-tuned for Concept Differentiation using SPCD. We compare them to their base pre-trained counterparts and to XL-LEXEME \citep{cassotti-etal-2023-xl}, a strong multilingual model trained on WiC-style data.

CALE achieve strong performance across evaluations: 79.3 balanced accuracy on Concept Differentiation, as well as top scores on Lexical Semantic Change Detection (a sense-level task) and in-context Lexical Similarity (a concept-level task). CALE models even match or outperform XL-LEXEME in non-English benchmarks, despite monolingual fine-tuning, demonstrating the strength of their fine-tuning objective and the base model XLM-R's cross-lingual capabilities.

Finally, analysis of the resulting embedding spaces reveals that CALE fine-tuning enables the shift from lemma-centric to concept-centric representations. We also show that CALE best reflects the conceptual structure of our reference lexical resources WordNet, with a correlation to similarities in WordNet above $.50$. 
We release both the SPCD dataset and the CALE models to support further research on token-level lexical meaning representations.

\section{Related Work}

\subsection{Lexical Semantic Tasks}

A variety of NLP tasks have explored the mapping between words and meanings, especially regarding polysemy and synonymy (or similarity in meaning in general). Word Sense Disambiguation (WSD) maps word occurrences to predefined senses \citep{navigli-2009-wsd, bevilacqua-etal-2021-wsd}, while its unsupervised counterpart, Word Sense Induction (WSI), clusters usages into latent sense groupings \citep{manandhar-etal-2010-semeval, jurgens-klapaftis-2013-semeval}. The Word-in-Context (WiC) task refines this by classifying whether two instances of a word share the same sense \citep{pilehvar-camacho-collados-2019-wic, martelli-etal-2021-semeval}. Lexical Semantic Change Detection (LSCD) extends WSI temporally, aiming to detect or rank meaning shifts across time periods \citep{schlechtweg-etal-2020-semeval, d-zamora-reina-etal-2022-black}. While these tasks focus on polysemy and intra-word variation, they do not address inter-word semantic similarities. Tasks such as Lexical Substitution \citep{mccarthy-navigli-2007-semeval, zhou-etal-2019-bert} and Lexical Similarity \citep{hill-etal-2015-simlex, vulic-etal-2020-multi, huang-etal-2012-improving, armendariz-etal-2020-cosimlex} examine inter-word relations, centering on synonymy and contextual interchangeability. Substitution tasks often involve ranking appropriate alternatives in context \citep{erk-pado-2008-structured, kremer-etal-2014-substitutes}.

More recently, Concept Induction has been proposed to unify word-level and cross-word perspectives by clustering usages of multiple target words into semantic concepts \citep{lietard-etal-2024-word}. Similarly, Frame Induction clusters word usages (typically verbs) into broader event frames with their associated arguments \citep{qasemizadeh-etal-2019-semeval, yamada-etal-2021-semantic, mosolova-etal-2024-injecting}. These tasks aim to capture higher-order abstractions that go beyond individual word senses or pairwise similarity.

\subsection{Embeddings for Semantic Representations}

Contextualized word embeddings from models like ELMo \citep{peters-etal-2018-deep} and BERT \citep{devlin-etal-2019-bert} generate dynamic representations based on usage in context, enabling better semantic modeling than earlier static or sparse methods. However, pooling token-level embeddings into meaningful sentence-level vectors remains challenging due to issues like anisotropy in the embedding space\footnote{i.e., the tendency for embeddings to occupy a narrow cone, leading to high cosine similarity for even unrelated items \citep{gao-2018-representation}.} \citep{reimers-gurevych-2019-sentence}. To improve semantic similarity and retrieval, Sentence-BERT (SBERT) \citep{reimers-gurevych-2019-sentence} introduced a Siamese training architecture, followed by advances using contrastive learning \citep{gao-etal-2021-simcse} and multi-task objectives \citep{zhang-etal-2022-task} to enhance cross-task generalization. Such models have been adapted for lexical semantics. \citet{mosolova-etal-2024-injecting} fine-tune language models using Wiktionary supervision to improve contextualized token embeddings for tasks like Word-in-Context. Similarly, the XL-LEXEME model \citep{cassotti-etal-2023-xl}, based on XLM-RoBERTa  \citep{conneau-etal-2020-unsupervised}, extends SBERT for multilingual word meaning tasks, showing robustness on LSCD benchmarks and supporting cross-lingual, context-sensitive embedding spaces.

Other efforts fine-tune BERT for word-level classification tasks \citep{gari-soler-apidianaki-2020-multisem}, though these cross-encoder models often lack flexibility for single-occurrence representation which represents a limitation for tasks requiring generalized concept modeling. Embedding-based models have also proven effective in capturing non-literal or stylistic nuances, as shown by applications in authorship verification \citep{rivera-soto-etal-2021-learning}. Additionally, SenseBERT \citep{levine-etal-2020-sensebert} enhances contextual embeddings by incorporating sense information from WordNet during pretraining, improving performance on sense-sensitive tasks and highlighting the benefit of explicitly modeling lexical semantics in transformer-based representations. Collectively, these developments demonstrate the adaptability of sentence-transformer architectures for fine-grained semantic representations.

\section{Concept Differentiation}
Our primary goal is to build a model able to accurately represent semantic information in order to distinguish concepts in context, from which we can obtain concept-aligned embeddings. To train such model, we rely on a supervised task that captures both polysemy and synonymy and propose Concept Differentiation, a supervised alternative to Concept Induction defined in \citet{lietard-etal-2024-word}. 

\subsection{Task Definition}
We define Concept Differentiation as a binary classification task where the goal is to determine whether two word usages represent the same underlying concept, regardless of whether they come from the same lemma or different lemmas. To illustrate, consider the following three phrases:
\begin{itemize}
    \item A. ``the boy could easily \underline{distinguish} the different note values''
    \item B. ``the patient’s ability to \underline{recognize} forms and shapes''
    \item C. ``the government had refused to \underline{recognize} their autonomy and existence as a state''
\end{itemize}

The verbs \textit{distinguish} in A and \textit{recognize} in B both express the concept of \textsc{discern} --- to tell things apart --- so this pair is labeled 1 (same concept). In contrast, while B and C both use \textit{recognize}, only C refers to \textsc{acknowledge}, so that pair is labeled 0 (different concepts). This task is broader than traditional Word-in-Context, which is limited to same-lemma comparisons. Here, we extend the comparison to any pair of word usages --- same lemma or not --- capturing finer and more flexible distinctions of meaning. It can also be seen as a binary reformulation of the Concept Induction task, where instead of clustering occurrences into concept clusters, we ask if two usages share a common concept. This setup allows us to train models that learn to compare usages in context, effectively capturing subtle semantic distinctions or consistencies between different instances of target words across the corpus.


\subsection{Dataset for Concept Differentiation}
In the absence of a dedicated dataset for this task, we construct a new dataset using the SemCor corpus, using its WordNet synsets annotations \citep{miller-1995-wordnet, fellbaum-1998-wordnet} as concept labels. For each lemma with enough labeled instances, we retrieve its occurrences following the procedure detailed in Appendix \ref{appdx:dataset}. We gather $70.3$k occurrences for $1902$ target lemmas, covering $5899$ concepts. To create labeled training pairs, we sample pairs of word occurrences $(\occany[j], \occany[k]) $ and assign a label of 1 if both instances are annotated with the same concept, and 0 otherwise, regardless of whether they are occurrences of the same target lemma or not.

To rigorously assess generalization, we enforce a strict separation between training, validation, and test sets by first partitioning the set of concepts and the set of target words into disjoint subsets for each split. Specifically, we randomly sample subsets of concepts and lemmas to be assigned exclusively to the validation or test split (5\% of concepts / lemmas in validation, 10\% in test). Once these partitions are defined, we extract all word occurrences from the SemCor corpus that are annotated with the selected concepts and lemmas, and include them in the corresponding data split. This ensures a more rigorous evaluation of a model’s ability to generalize to entirely unseen meanings and lexical items, preventing models from relying on memorized associations between words and concepts. In each split separately, we create pairs by attempting to match each occurrence to 4 others, one for each of the following categories if possible: same concept and same lemma (SC/SL), same concept but different lemma (SC/DL, only if we can find an occurrence of another lemma for the concept), different concept but same lemma (DC/SL, only if the word is polysemous) and different concept and different lemma (DC/DL, unrelated pairs). The dataset contains $44$k pairs (from the $14.3$k reserved occurrences) in the \textit{test} split, $20$k (from the $6.5$k reserved occurrences) in the \textit{validation} split, and $156$k (from the $49.6$k remaining occurrences) in the \textit{train} split. In all splits, the resulting proportion of label `1' (same concept) is around $40$\%.

The resulting dataset, later referred as ``\underline{S}emCor \underline{P}airs for \underline{C}oncept \underline{D}ifferentiation'' (SPCD), supports research in semantic representation and disambiguation and is made publicly available to facilitate future benchmarking and exploration\footnote{\url{https://hf.co/datasets/gabrielloiseau/CALE-SPCD/}}.  Further description as well as more details about the extraction process can be found in Appendix \ref{appdx:dataset}.

\section{Concept-Aligned Embeddings}
We use our resulting pair classification dataset to train a representation model. The goal is to learn a function $f$ mapping an occurrence $\occany[i]$ to a vector representation $\embany[i]$ in a vector space in which the embeddings of two occurrences of the same concept have higher cosine similarity than embeddings of two occurrences of a different concept. 
To fit $f$, we adopt a Siamese representation learning architecture, inspired from SentenceBERT. Concretely, our model consists of a pre-trained contextualized language model followed by a pooling layer that produces a fixed-dimensional vector for an input sequence. During training, two identical copies of this encoder sharing the same parameters are used to process two occurrences as a training pair. The parameters of the encoder are updated jointly, ensuring that both occurrences are embedded in the same representation space.

Each training step considers a batch of occurrence pairs $(\occany[i], \occany[j])$ with an associated label $ y_{ij} \in \{0, 1\} $, where $ y_{ij} = 1 $ indicates that both occurrences share the same semantic concept, and $ y_{ij} = 0 $ otherwise. Occurrence of the pair are encoded into vector representations $(\embany[i], \embany[j])$ by the Siamese models (two instances of the same model with shared weights during training). We then compute the cosine similarity between the two embeddings $ \mathrm{sim}_{ij} = \cos(\embany[i], \embany[j])$.
The training objective is to maximize similarity for positive pairs ($ y_{ij} = 1 $) and minimize it for negative pairs ($ y_{ij} = 0 $). We fit the model to minimize a contrastive loss \cite{contrastiveloss} defined as:
$$
\frac{1}{2} \left[ y \cdot \operatorname{sim}_{ij}^2 + (1 - y) \cdot \max(0, m - \operatorname{sim}_{ij})^2 \right]
$$
where $ m $ defines the minimum distance that should exist between dissimilar (negative) pairs in the embedding space, known as the margin.

This contrastive training encourages the model to organize the embedding space such that occurrences sharing the same concept are pulled closer together, while unrelated occurrences are pushed apart. Fine-tuning under this objective enables the model to produce embeddings that better reflect conceptual similarity across varied lexical and contextual realizations.

An occurrence $\occany[i]$ is a sequence of words $(t_1, \ldots, t_n)$ with a target word $w$ in position $k$ ($t_k=w$). To be used as input for the model, the occurrence is transformed as follows: 
$$
    \underbrace{t_1, \ldots, t_{k-1}}_{\text{left context}}, \text{<t>}, w, \text{</t>}, \underbrace{t_{k+1} \ldots, t_n}_{\text{right context}},
$$
using special tags <t></t> to delimit the target word in the sentence. During training, the model will learn that the contrast must be made based on the meaning of words between the delimiters. In doing so, we ensure the embedding corresponding to the full sequence is the representation of $\occ{i}$ and distinct from the embedding of another target word occurring in the same sentence, without needing to account for potential subwords that may result from the model's tokenizer. This method, also used by \citet{cassotti-etal-2023-xl}, makes the model easy to use in practice at inference time, as we only have to add/move the delimiters to shift the model's focus from one word to another in the sentence. 

We refer to this fine-tuning approach and the resulting models as \textbf{CALE} (\underline{C}oncept \underline{Al}igned \underline{E}mbeddings) and adopt this terminology throughout the paper. Our models are made publicly available through HuggingFace\footnote{\href{https://hf.co/collections/gabrielloiseau/cale-689344d272fb2b658f881c28}{[Hyperlink to HF-models]}}. We limited the scope of our study to single-word targets, but our framework could be applied to capture the meaning of Multi-Word Expressions in future work.

\section{Model Evaluation}

In this section, we report the performances of different models, including models with CALE fine-tuning, for the Concept Differentiation task on the test split of the SPCD dataset, as well as other external lexical semantic tasks, namely Lexical Semantic Change Detection and In-context Lexical Similarity.

\paragraph{Compared Models.}
XLM-R \citep{conneau-etal-2020-unsupervised} is based on the RoBERTa Large architecture and pretrained on cross-lingual data. 
XL-LEXEME \citep{cassotti-etal-2023-xl} is a XLM-R model fine-tuned for multilingual WiC. Compared to our models, XL-LEXEME's fine-tuning was rather data-hungry, as three Word-in-Context datasets were used. XL-LEXEME established a new State-of-the-Art in some semantic tasks and is therefore our main reference point. Regrettably, models from \citet{gari-soler-apidianaki-2020-multisem} and \citet{mosolova-etal-2024-injecting} are not publicly available, preventing us to include them in our experiments without re-implementing the whole dataset curation process and fine-tuning method.
Our CALE models generate embeddings in a similar fashion than XL-LEXEME, but they are fine-tuned on Concept Differentiation, as described earlier. We experiment with several base model: XLM-R (to compare with XL-LEXEME) XL-LEXEME (to observe the impact of double fine-tuning and of XL-LEXEME's fine-tuning multilingually) and a monolingual English ModernBERT \citep{warner-etal-2024-modernbert}, a more-recent State-of-The-Art actualization of the BERT model, evaluated only in English datasets in order to discuss the potential difference between language specialization and pre-training multilingually.

\paragraph{Embedding extraction.}\,
For XLM-R, embeddings of word occurrences are obtained by averaging first over a specified range of layers, and then over subword tokens that compose the target word of the occurrence.
XL-LEXEME and CALE models are SentenceBERT models, meaning that a single embedding is produced per occurrence using a final pooling layer that averages the last layer embeddings of all tokens in the sequence. In that case, the delimiters around the target word ensure that the representation is specific to that target.

\paragraph{Hyperparameters and metrics.}\,
For XLM-R, we test three layer ranges. The \texttt{mid} layer set corresponds to intermediate-high layers (14 to 17), following the findings of \citet{chronis-erk-2020-bishop}; the \texttt{last} set (layers 21 to 24) is in line with a large number of works in lexical semantic tasks using the average over the last four layers as input; and \texttt{first} (layers 1 to 4) is included for comprehensiveness.
For all models and across all experiments, cosine distance is used as the standard metric of the representational space to compare two embeddings.

\subsection{Concept Differentiation Evaluation}

In Table \ref{tab:accuracies} we present the results for the Concept Differentiation task on the test split of the SPCD datasets. 

\paragraph{Threshold-based classifiers.}\,
For each candidate model, the classification process for a given pair of occurrences is the following: 
we compute the embeddings for each occurrence using the candidate model, then compare the cosine distance between them to a threshold to label the pair: $1$ (same concept) if the cosine distance below the threshold, $0$ (different concepts) otherwise. The threshold is chosen to maximize accuracy on the train and validation splits.

\paragraph{Baseline.}\,
To provide a baseline system, we label a pair as $1$ if and only if the two occurrences are from the same lemma, and $0$ otherwise. We refer to this baseline as \texttt{1L1C} for ``1 Lemma 1 Concept''. 

\paragraph{Evaluation metrics.}\,
We use Balanced Accuracy (the average of the Recalls from both classes) as the classes are not exactly balanced. We also tried F1 scores and the observed tendencies are the same, thus we only report them in Appendix \ref{appdx:cdiff-results}, with also a more complete overview of accuracies per class.

\begin{table}[t]
    \centering
    
    \begin{tabular}{lccc}\toprule
        & \textbf{All} & \textbf{SL} & \textbf{DL}\\
        \midrule\arrayrulecolor{white}
        1L1C (baseline) & 65.1 & 50.0 & 50.0  \\
        \midrule
        XLM-R (first) & 60.8 & 54.2 & 50.0 \\
        XLM-R (mid) & 70.9 & 59.1 & 68.4  \\
        XLM-R (last) & 69.1 & 59.6 & 63.5  \\
        XL-LEXEME & 76.7 & 62.4 & 82.7 \\
        \midrule
        CALE (XLM-R) & \textbf{79.3} & 65.9 & \textbf{86.4} \\
        CALE (MBERT) & 78.7 & 66.6 & 83.5 \\
        CALE (XL-LEX) & 79.2 & \textbf{67.2} & 84.4  \\
        \arrayrulecolor{black}\bottomrule
\end{tabular}
    \caption{Balanced Accuracies on test pairs from SPCD. ``SL'' (resp. ``DL'') to pairs of occurrences of the same lemma (resp. of different lemmas).}
    \label{tab:accuracies}
    \vspace{-0.8em}
\end{table}

\paragraph{Discussion.}\,
Our first observation is that the 1L1C baseline reaches high accuracy, providing a challenging baseline. This is in line with prior observations in the context of clustering \citep{lietard-etal-2024-word}. This indicates that the mapping between words and concepts in SemCor may resemble a one-to-one mapping with some deviations, and as such a model for Concept Differentiation would benefit from not ignoring whether two occurrences are of the same lemma or not.
We also observe that XLM-R performs best using late-intermediate layers than other layers and outperforms the baseline verifying findings of \citet{chronis-erk-2020-bishop} that these layers are best for tasks related to semantic similarity.

We also find that XL-LEXEME largely outperforms XLM-R even when presented pairs of Different Lemmas, meaning that by learning to differentiate senses of the same lemma, it captured more global information about relations between different words.

We find that CALE models outperforms XL-LEXEME. It is expected to some extent, as Concept Differentiation is the task CALE has been fine-tuned for. We also observe that Monolingual CALE (MBERT) is not better than the multilingual ones, underlining the advantage of multilingual pre-training.

The CALE model based on XL-LEXEME is globally on-par with the one based on XLM-R (79.3 and 79.2 balanced accuracies). The former is slightly better in the Same-Lemma setting (i.e. better at distinguishing senses) than the latter, but slightly worse in the Different-Lemma setting (i.e. worse at identifying synonyms), as expected because XL-LEXEME has been trained strictly in the Same-Lemma setting and therefore likely provides benefit in this context only.

\subsection{Lexical Semantic Change Detection.}

To assess models’ ability to capture meaning variation within lemmas across contexts, we evaluate on the Lexical Semantic Change Detection (LSCD) task, following ``Subtask 2`` of \citet{schlechtweg-etal-2020-semeval}. We use Diachronic Word Usage Graphs (DWUGs) datasets across six languages: English, German, Latin, Swedish, Spanish, and Italian \citep{schlechtweg-etal-2020-semeval, d-zamora-reina-etal-2022-black, cassotti-etal-2024-dwugs}\footnote{\url{https://www.ims.uni-stuttgart.de/en/research/resources/experiment-data/wugs/}}. Each dataset contains target words with occurrences sampled from two time periods ($T_1$, $T_2$), and the task is to assign a change score reflecting how much a word’s meaning has shifted. LSCD is a intra-lemma task: only usages of the same word are compared and semantic change is assessed by measuring differences in the word's senses between time periods. Performance is measured by Spearman correlation between predicted and gold change scores.
Note that \citet{cassotti-etal-2023-xl} also evaluated their model XL-LEXEME on these benchmarks and established new State-of-the-Art scores in English, German and Swedish.

We use the Average Pairwise Divergence (APD) measure from \citet{kutuzov-giulianelli-2020-uio} to compute change scores, relying on cosine distances between embeddings from $T_1$ and $T_2$:
$$
\operatorname{APD}(w) = \frac{1}{n\times m} \sum_{i=1}^n\sum_{j=1}^m d(\embT{i}{1}, \embT{j}{2})
$$
with $d$ the cosine distance operator, $\embT[w]{i}{k}$ the embedding of the $i$-th occurrence of word $w$ at time $T_k$, for a target word $w$ with $n$ occurrences at $T_1$ and $m$ occurrences at $T_2$. 
APD is generally accepted as standard for Lexical Semantic Change (see the survey study of \citet{periti-montanelli-2024-survey}) and was used by \citet{cassotti-etal-2023-xl} to evaluate XL-LEXEME on the same task on DWUGs datasets.
In Appendix \ref{appdx:dwug-results} we tried PRT, another measure from \citet{kutuzov-giulianelli-2020-uio}.

\begin{table}[t]
    \centering
    \resizebox{\columnwidth}{!}{
    \begin{tabular}{lcccccc}
        \toprule
         & EN & DE & ES & IT & LA & SV \\
        \arrayrulecolor{white}\midrule\arrayrulecolor{black}
        \# targets & 46 & 50 & 100 & 26 & 40 & 44 \\
        \midrule\arrayrulecolor{white}
        XLM-R (first) & .22 & .14 & .08 & .18 & .00$^\star$ & .10 \\
        XLM-R (mid) & .57 & .54 & .49 & .32 & \textbf{.24}$^\star$ & .58 \\
        XLM-R (last) & .56 & .34 & .53 & .27 & .00$^\star$ & .52 \\
        XL-LEXEME & \textbf{.79} & .80$^\star$ & .63$^\star$ & .51$^\star$ & .14$^\star$ & .85$^\star$ \\
        \midrule
        CALE (XLM-R) & .71$^\star$ & .78$^\star$ & .65$^\star$ & \textbf{.75} & .11$^\star$ & .83$^\star$ \\
        CALE (MBERT) & .66$^\star$ & - & - & - & - & - \\
        CALE (XL-LEX) & .73$^\star$ & \textbf{.83} & \textbf{.68} & .54 & .20$^\star$ & \textbf{.86} \\
        \arrayrulecolor{black}\bottomrule
    \end{tabular}}
    \caption{Spearman Correlation between model Average Pairwise Divergence (APD) and gold change scores in DWUG datasets. The highest score for each dataset is in bold, and other scores that are not significantly different from it are marked with $^\star$.}
    \label{tab:lscd-apd}
    \vspace{-1.5em}
\end{table}

We report APD-based results in Table~\ref{tab:lscd-apd}. We use Steiger's Z-test on dependent correlations to determine whether differences to the best-performing model are significant or not \citep{steiger-1980-tests}.

Results show that CALE fine-tuning consistently improves over base XLM-R in all languages, even though it was trained on English data only, showcasing the cross-lingual transferability of XLM-R. CALE also matches or slightly outperforms XL-LEXEME in 5 out of 6 languages, despite XL-LEXEME being fine-tuned on larger multilingual datasets. As in prior work, Latin proves challenging across all models.

While XL-LEXEME is fine-tuned for Word-in-Context (a lemma-centric task closely related to LSCD), CALE is optimized for Concept Differentiation, a broader task involving both intra- and inter-lemma comparisons. One might expect this to hinder LSCD performance, yet results suggest otherwise: CALE (XL-LEXEME) slightly improves over XL-LEXEME alone, suggesting that Concept Differentiation enhances a model’s ability to detect semantic change. Due to the limited number of target words, differences are not statistically significant, but findings indicate that CALE fine-tuning does not degrade LSCD performance.

\subsection{Contextual Lexical Similarity.}

\begin{table}[t]
    \centering
    \begin{adjustbox}{max width = \columnwidth}
    \begin{tabular}{p{0.8\columnwidth}c}
    \toprule
    \textbf{Context} & \textbf{Similarity} \\
    \midrule
         $[...]$ As J.T. \underline{put} it, if Kaz can't follow orders, he might as well learn to \underline{give} them. & 2.88 \\
         $[...]$ Caleb returns to the table to let Kirsten know that he will \underline{give} his support to Julie at the board meeting. $[...]$ but for her marriage's sake, she will \underline{put} her support to Sandy. & 7.83 \\
    \bottomrule
    \end{tabular}
    \end{adjustbox}
    \caption{Example entry of CoSimLex context pairs with the mean similarity rating between target words (underlined).}
    \label{tab:cosimlex_examples}
\end{table}

\begin{table}[t]
    \centering
    \begin{adjustbox}{max width=\columnwidth}
    \begin{tabular}{lcccc}
        \toprule
        \multicolumn{5}{c}{\large \textsc{\bfseries Subtask 1 : Contextual Changes}}\\\midrule
         & EN & FI & HR & SL \\
        \arrayrulecolor{white}\midrule\arrayrulecolor{black}
        \# examples & 340 & 24 & 112 & 111 \\
        \midrule\arrayrulecolor{white}
        XLM-R (first) & .18 & .32$^\dagger$ & .20 & .33 \\
        XLM-R (mid) & .69$^\star$ & .36$^\dagger$ & .58 & .52 \\
        XLM-R (last) & .63 & .27$^\dagger$ & .51 & .45 \\
        XL-LEXEME & .70 & .47$^\star$ & .67$^\star$ & .66$^\star$ \\
        \midrule
        CALE (XLM-R) & .74$^\star$ & \textbf{.63} & \textbf{.72} & \textbf{.70} \\
        CALE (MBERT) & .68 & - & - & - \\
        CALE (XL-LEX) & \textbf{.74} & .60$^\star$ & .72$^\star$ & .67$^\star$ \\
        \arrayrulecolor{black}\bottomrule \\
        \toprule
        \multicolumn{5}{c}{\large \textsc{\bfseries Subtask 2 : Similarity Ratings}}\\\midrule
         & EN & FI & HR & SL \\
        \arrayrulecolor{white}\midrule\arrayrulecolor{black}
        \# pairs & 680 & 48 & 224 & 222 \\
        \midrule\arrayrulecolor{white}
        XLM-R (first) & .27 & -.04$^\dagger$ & .25 & .25 \\
        XLM-R (mid) & .61 & .25$^\dagger$ & .53 & .51 \\
        XLM-R (last) & .59 & .16$^\dagger$ & .52 & .46 \\
        XL-LEXEME & .63 & .59$^\star$ & .64 & .67$^\star$ \\
        \midrule
        CALE (XLM-R) & \textbf{.67} & \textbf{.64} & .68$^\star$ & \textbf{.68} \\
        CALE (MBERT) & .67$^\star$ & - & - & - \\
        CALE (XL-LEX) & .67$^\star$ & .57$^\star$ & \textbf{.73} & .68$^\star$ \\
        \arrayrulecolor{black}\bottomrule
    \end{tabular}
    \end{adjustbox}
    \caption{Pearson Correlation of CoSimLex Subtask 1 (upper table) and Spearman Correlation of CoSimLex Subtask 2 (lower table). Values marked with $^\dagger$ are non-significant correlation values. The highest score for each dataset is in bold, and other scores that are not significantly different from it are marked with $^\star$.}
    \label{tab:cosimlex-scores}
    \vspace{-1em}
\end{table}

We now evaluate candidate models in the cross-lemma setting.
\paragraph{Dataset.}\,
To assess our models' ability to represent conceptual similarity of two distinct words in a given context, we use the CoSimLex dataset \citep{armendariz-etal-2020-cosimlex} for In-Context Lexical Similarity. This resource proposes two subtasks in 4 different languages, English, Finnish, Croatian and Slovene. Each entry of CoSimLex is a pair of target words, and two different contexts in which both words appear. For each context is provided a measure of lexical similarity between the two words. An example of a CoSimLex entry in English is provided in Table \ref{tab:cosimlex_examples}, where the respective meanings of \textit{put} and \textit{give} are closer in context 2 than in context 1.

\paragraph{Tasks.}\,
Subtask 1, \textit{Contextual Changes}, aims at predicting the change of similarity between the two words, from context 1 to context 2. This is measured with Pearson Correlation coefficient between the differences of gold similarities and the differences of predicted similarities. Subtask 2, \textit{Similarity Ratings}, adopts a more traditional setting, where candidate systems are simply tasked to predict a measure of lexical similarity between the two words in each context, and measuring the Spearman Correlation between gold measures of similarity and predicted ones. Subtask 2 uses independently each context provided in Subtask 1 instead of pairing them, and as such covers twice as many examples. 

\paragraph{Predicting lexical similarity in the same-context setting.}\,
For both subtasks, given a context from CoSimLex where two target words appear, we compute the embeddings corresponding to each target and then compare them using cosine distance.
For XLM-R, we take the embeddings corresponding to the two sets of subwords, that compose each target. 
For XL-LEXEME and CALE, we create two separate input sequences: the first input contains the full context with our delimiter tags placed around the first target word only; and the second with delimiters around the second target only.
Feeding each sequence to the model, we obtain two embeddings, each representing the meaning of the specifically tagged target word.

\paragraph{Results and discussion.}\,
Results are displayed in Table \ref{tab:cosimlex-scores}.
CALE models reach the highest correlations values in all languages, ranging between $0.63$ to $0.73$ in subtask 1 and $0.57$ to $0.68$ in subtask 2, and consistently improve over XLM-R. Surprisingly, XL-LEXEME also performs very well and is most often not deemed significantly different from CALE results, despite being trained only on a same-lemma task and not on inter-words relations. CALE (XL-LEX) does not perform significantly better than the more simple CALE (XLM-R), indicating that CALE fine-tuning is usually enough to obtain accurate in-context similarities. Again, the good performances reached by CALE fine-tuning highlight the cross-lingual transferability of XLM-R models.

\section{Spatial Organization}

\begin{table}
    \centering
    \resizebox{\columnwidth}{!}{
    \begin{tabular}{lccc}
        \toprule
         & Silh (S) & Silh (UB) & WuP $\rho$ \\
        \midrule\arrayrulecolor{white}
        XLM-R (first) & -0.166 & -0.267 & 0.187 \\
        XLM-R (mid) & -0.064 & -0.261 & 0.338 \\
        XLM-R (last) & -0.088 & -0.260 & 0.311 \\
        XL-LEXEME & -0.124 & -0.370 & 0.489 \\
        \midrule
        CALE (XLM-R) & -0.041 & -0.192 & 0.509 \\
        CALE (MBERT) & \textbf{0.042} & \textbf{-0.104} & 0.514 \\
        CALE (XL-LEX) & 0.003 & -0.228 & \textbf{0.532} \\
        \arrayrulecolor{black}\bottomrule
    \end{tabular}}
    \caption{Silhouette scores with synsets (S) or with Unique Beginners (UB) as cluster labels ; and Spearman Correlation of cosine similarities with Wu-Palmer's similarities in WordNet.}
    \label{tab:geometry}
    \vspace{-1.5em}
\end{table}

\begin{figure}[ht]
    \centering
    \includegraphics[width=\linewidth]{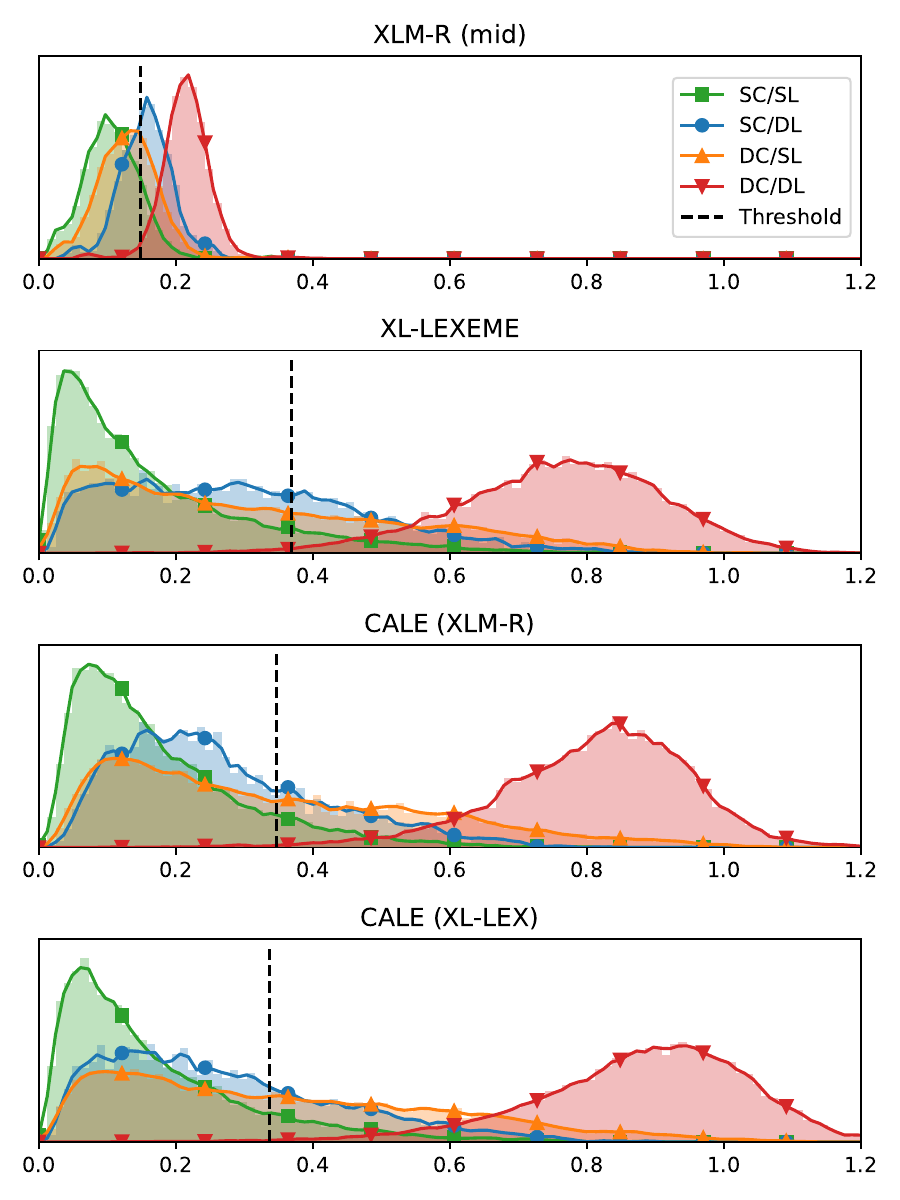}
    \caption{Distribution of cosine distance between embeddings from the candidate models' spaces for SPCD \textit{test} pairs, according to their category. The dashed line is the decision threshold for Concept Differentiation learned from the other splits.}
    \label{fig:distrib}
    \vspace{-1.5em}
\end{figure}

We investigate how CALE fine-tuning affects representation space structure. Figure \ref{fig:distrib} shows cosine distance distributions across the \textit{test} split of SPCD by pair category, while Table~\ref{tab:geometry} analyzes how well embedding spaces reflect WordNet's conceptual organization using synsets annotations.

\paragraph{Distance distributions.} In Figure \ref{fig:distrib} we plot the distributions of distance in pairs according to their categories: `SL' and `DL' refer to ``same lemma'' and ``different lemmas'' respectively, `SC' and `DC' to ``same concept'' and ``different concepts'' respectively. We first observe that XLM-R suffers from major anisotropy (i.e. all cosine distances are low even in unrelated pairs of the DC/DL category). The fine-tuning of CALE and XL-LEXEME pushes occurrences in DC/DL pairs away from one another, a consequence of the margin parameter. Another observation is that CALE's fine-tuning successfully shifts the focus from \textit{lemma}-centric to \textit{concept}-centric: in \texttt{XLM-R}, DC/SL pairs (cases of polysemy) are generally closer than SC/DL pairs (cases of synonymy); this tendency is reversed in \texttt{CALE (XLM-R)} (average distance of $.130$ for DC/SL vs $.157$ for SC/DL before fine-tuning; $.343$  vs $.269$ after fine-tuning).
Yet, we remark that the cosine distances for several DC/SL pairs (examples of the polysemy of a single lemma) remain low: distances in the DC/SL category are not distributed in a tight mass, but rather a large band across the cosine distance range.
We hypothesize that the reason for this low cosine distance value is that this corresponds to pairs of similar senses.
Senses of a given lemma can be very similar to one another, and WordNet is known to be very fine-grained in its sense distinctions.
Indeed, after examining this phenomenon more closely, we have found a (significant) correlation coefficient of 0.2 between Wu-Palmer similarity in WordNet and cosine distances between CALE (XL-LEX) embeddings. Thus, surprisingly low cosine distances in DC/SL pairs may be explained by similarity between word senses in the reference resource directly (at least partially, given that the correlation value is significant but not very high).

\paragraph{Similarity correlations.}
WordNet's synsets are linked to one another through relations of hyper-/hyponymy, in a tree-like structure. A short path between two synsets indicate that the two concepts are similar. We compute Spearman correlations between model's cosine similarities and Wu-Palmer similarities \citep{wu-palmer-1994-verb} between synsets for each occurrence pair\footnote{We also tried Lin similarity, with similar results.}. CALE models achieve the highest correlations (above 0.5), indicating a strong alignment between embedding similarity and semantic similarity. All correlations and pairwise differences are found statistically significant.

\paragraph{Cluster analysis.}
We evaluate whether occurrences of the same concept form tight, well-separated clusters using the Silhouette score \citep{ROUSSEEUW198753}, with WordNet synsets as labels. Silhouette score measures how easily occurrences would be assigned to their cluster by comparing the distance to the occurrence's cluster to the distance to the closest other cluster. A score close to 1 indicates that cluster are well-formed and well-separated. Scores close to 0 indicate that there exist overlaps between clusters, and scores close to -1 indicate that occurrences are almost systematically assigned to the wrong cluster. CALE models yield the highest scores, suggesting better conceptual clustering. However, all scores remain below 0.1, highlighting substantial overlap between clusters even for the best models, which is likely due to the fine granularity of WordNet synsets.

\paragraph{Broader clusters.}
We repeat the clustering analysis using WordNet’s Unique Beginners (UB) broad semantic categories like ``entity'', ``feeling'', or ``communication'' as labels, limited to nouns and verbs. Overall scores are negative, indicating poor alignment with these higher-level groupings. Nonetheless, CALE performs best among models, suggesting a modest ability to reflect coarse semantic distinctions.

\paragraph{Conclusions.}
These findings show that CALE embeddings align more closely with WordNet’s structure than other models, both in distances and clustering. 
Notably, ModernBERT-based CALE shows stronger structural alignment than multilingual models. XL-LEXEME, despite good pairwise similarity, shows weaker clustering, presumably because its fine-tuning was sense-level and did not require broad conceptual organization.

\section{Conclusion}

We introduced Concept Differentiation, a supervised task for predicting whether two occurrences share the same semantic concept. We curated the SPCD dataset from SemCor and developed CALE models using contrastive fine-tuning of SentenceBERT-style architectures.

Evaluation shows that CALE improves lexical semantic representations across tasks, including Concept Differentiation, Lexical Semantic Change Detection, and Lexical Similarity. These gains hold for both intra-lemma and inter-lemma tasks, with CALE matching or surpassing strong baselines like XL-LEXEME, even on multilingual benchmarks despite English-only training. This suggests that Concept Differentiation subsumes other lexical tasks dealing with the mapping between words and meanings, and that a broader view is beneficial for multi-purpose semantic representations.

Analysis of CALE's embedding geometry reveals a shift from lemma-centric to concept-centric representations and closer alignment with WordNet's organization.  These results position Concept Differentiation as a promising general-purpose training signal for lexical semantics.

Future work could explore specifically-curated datasets for Concept Differentiation, with human annotators and a different annotation scheme than WordNet's synsets. We also look forward to applications of CALE to Multiword Expressions (MWE), and potential applications to Frame Induction, Concept Induction or other semantic tasks.

\section*{Limitations}
While CALE models demonstrate cross-lingual transferability, fine-tuning was performed exclusively on English language data. Even if multilingual models (e.g., XLM-R) generalize well to other languages in tasks like LSCD and CoSimLex, the performance in lower-resource or typologically distant languages has not been thoroughly evaluated due to the lack of data sources. A more rigorous multilingual training and evaluation setup is needed to assess CALE's robustness across languages.

Second, the binary nature of Concept Differentiation may oversimplify semantic similarity, which is often graded rather than categorical. This could limit the model’s applicability in tasks requiring nuanced semantic judgments.

Third, while we show improved alignment with WordNet's conceptual structure, our models receive no direct supervision about ontological hierarchy or fine-grained relations between concepts (e.g., hyper-/hyponymy, holo-/meronymy, antonymy). Further work is needed to evaluate and potentially incorporate structured knowledge explicitly during training.

Lastly, we relied on small architectural modifications compared to XL-LEXEME for model comparison, a key aspect of our studies is model and data scaling. However, other scaling axes, notably in terms of model parameters and embedding dimension size are left unexplored.

\section*{Acknowledgments}

We gratefully thank the anonymous reviewers for their insightful comments.
This research was funded by Inria Exploratory Action COMANCHE.
We would like to thank Pascal Denis and Mikaela Keller for their valuable feedback on the early draft of this paper.

\bibliography{custom, anthology}

\appendix

\section{SPCD Dataset}
\label{appdx:dataset}

To extract occurrences from the SemCor corpus \citet{fellbaum-1998-wordnet}, we proceeded as follows:
\begin{itemize}
    \item We used SemCor as distributed through the NLTK\footnote{\url{https://www.nltk.org/install.html}} library.
    \item We only used SemCor sentences that had semantic annotations (of WordNet's synsets) and contained minimum 10 words and no more than 100.
    \item For each sentence, we extract an occurrence for each tagged word in the sentence (retrieved occurrences are delimited by the sentence).
    \item We focus on Adjective, Nouns and Verbs, using the Part-of-Speech tag indicated by the WordNet's synsets. We merge PoS-tags ``a'' and ``s'' as they both cover Adjectives.
    \item For each PoS independently (to filter-out potential noisy PoS annotations), we keep only lemmas that meet the following conditions:\begin{itemize}
        \item has minimum 3 letters.
        \item occurs at least 10 times with the PoS.
        \item is a single word (no compound) and contain only letters (no symbol/number).
        \item is not a proper noun.
    \end{itemize}
    \item We discard all occurrences of non-selected lemmas for each PoS tag, and then join the 3 remaining sets of lemmas and their occurrences. Now that extraction and filtering are done, we no longer differentiate on PoS tags (we allow pairs of occurrences from different PoS tags).
\end{itemize}

Table \ref{tab:dataset} provides the number of pairs in each split and category.

\section{Fine-Tuning Hyperparameters}
\label{appdx:hyperparams}
We conducted hyperparameter optimization using Optuna's\footnote{\url{https://optuna.org/}} Tree-structured Parzen Estimator (TPE) \cite{NIPS2011_86e8f7ab} algorithm to identify the best-performing configuration for our model. The optimal hyperparameters found through this search are presented in Table \ref{tab:hyperparams}.

\begin{table}[h]
    \centering
    \begin{tabular}{lc}
        \toprule
        Hyper-parameter & Value \\
        \midrule
        Margin & 0.7 \\
        Learning Rate & 6.02e-06 \\
        Warmup Ratio & 0.24 \\
        Weight Decay & 0.05 \\
        Adam $\beta_1$ & 0.9 \\
        Adam $\beta_2$ & 0.999 \\
        Adam $\epsilon$ & 1e-08 \\
        Epochs & 1 \\
        Seed & 42 \\
        Embedding Size & 1024 \\
        \bottomrule
    \end{tabular}
    \caption{CALE models hyper-parameters.}
    \label{tab:hyperparams}
\end{table}

\section{Test Metrics in SPCD}
\label{appdx:cdiff-results}
Table~\ref{tab:accuracies_full} presents more detailed evaluation results on the SPCD dataset, distinguishing between the different pair categories (Same concept/Different concepts and Same lemma/Different lemma). 

Table~\ref{tab:f1} shows the F1 score in the \textit{test} split of SPCD.

\begin{table}[ht]
    \centering
    \begin{tabular}{lccc}\toprule
             & \textbf{All} & \textbf{SL} & \textbf{DL} \\\midrule\arrayrulecolor{white}
            1L1C (baseline) & 62.8 & 69.5 & 0.0 \\\midrule
            XLM-R (first) & 50.2 & 57.8 & 0.0 \\
            XLM-R (mid) & 67.1 & 69.7 & 53.3 \\
            XLM-R (last) & 64.2 & 68.2 & 42.8 \\
            XL-LEXEME & 73.5 & 72.5 & 78.2 \\\midrule
            CALE (XLM-R) & \textbf{76.2} & 74.5 & \textbf{83.6} \\
            CALE (MBERT) & 75.4 & 74.4 & 79.9 \\
            CALE (XL-LEX) & 76.0 & \textbf{74.8} & 81.2 \\
            \arrayrulecolor{black}\bottomrule
    \end{tabular}
    \caption{F1 scores on test pairs from SemCor. ``SL'' (resp. ``DL'') refers to pairs of occurrences of the same lemma (resp. of different lemmas) and ``SC'' (resp. ``DC'') to pairs of occurrences referring to the same concept (resp. to different concepts).}
    \label{tab:f1}
\end{table}

\section{APD and PRT results on DWUGs}
\label{appdx:dwug-results}

In addition to APD, we also tried another measure from \citet{kutuzov-giulianelli-2020-uio}, the Prototype Distance (PRT), defined as follows:

$$
\operatorname{PRT}(w) = d( \frac{1}{n}\sum_{i=1}^n\embT{i}{1} \;,\; \frac{1}{m}\sum_{j=1}^m\embT{j}{2})
$$
with $d$ the cosine distance operator, $\embT[w]{i}{k}$ the embedding of the $i$-th occurrence of word $w$ at time $T_k$, for a target word $w$ with $n$ occurrences at $T_1$ and $m$ occurrences at $T_2$. 

Side-by-side results of APD and PRT can be found in Table \ref{tab:dwugs-complete}.

\begin{table*}[h]
    \centering
    \resizebox{\textwidth}{!}{
    \begin{tabular}{lccccccccc}
        \toprule
          & \textbf{Total}& \multicolumn{2}{c}{\textbf{Concept}} & \multicolumn{2}{c}{\textbf{Lemma}} & \multicolumn{4}{c}{\textbf{Cross-categories}}\\
        \cmidrule(lr){3-4}
        \cmidrule(lr){5-6}
        \cmidrule(lr){7-10}

        \textbf{Split} &  & \textbf{Same} & \textbf{Different} & \textbf{Same} & \textbf{Different} & \textbf{SC\&SL} & \textbf{SC\&DL} & \textbf{DC\&SL} & \textbf{DC\&DL}\\
        \midrule
        Train & 156,387 & 63,752 \textit{(41\%)} & 92,635 \textit{(59\%)} & 91,551 & 64,836 & 48,517 & 15,235 & 43,034 & 49,601 \\
        Val & 20,346 & 8,114 \textit{(40\%)} & 12,232 \textit{(60\%)} & 12,115 & 8,231 & 6,387 & 1,727 & 5,728 & 6,504 \\
        Test & 44,891 & 18,317 \textit{(41\%)} & 26,574 \textit{(59\%)} & 26,318 & 18,573 & 14,018 & 4,299 & 12,300 & 14,274 \\
        \bottomrule
    \end{tabular}
    }
    \caption{SPCD dataset description in number of pairs. The train split used 49,601 unique occurrences, 6,504 for validation and 14,274  for test.``SL'' (resp. ``DL'') refers to pairs of occurrences of the same lemma (resp. of different lemmas) and ``SC'' (resp. ``DC'') to pairs of occurrences referring to the same concept (resp. to different concepts).}
    \label{tab:dataset}
    \vspace{1em}
\end{table*}

\begin{table*}[h]
    \centering
    \begin{tabular}{lccccccc}\toprule
         & \textbf{All}$^\dagger$ & \textbf{SL}$^\dagger$ & \textbf{DL}$^\dagger$ & \textbf{SC\&SL} & \textbf{DC\&SL} & \textbf{DC\&DL} & \textbf{SC\&DL} \\
         \midrule\arrayrulecolor{white} 
        1L1C (baseline) & 65.1 & 50.0 & 50.0 & 100 & 0.0 & 100 & 0.0 \\\midrule
        XLM-R (first) & 60.8 & 54.2 & 50.0 & 54.1 & 51.5 & 100 & 0.0 \\
        XLM-R (mid) & 70.9 & 59.1 & 68.4 & 84.6 & 30.4 & 95.5 & 39.5 \\
        XLM-R (last) & 69.1 & 59.6 & 63.5 & 77.9 & 38.4 & 95.8 & 28.8 \\
        XL-LEXEME & 76.7 & 62.4 & 82.7 & 89.7 & 35.0 & 98.2 & 67.4 \\\midrule
        CALE (XLM-R) & 79.3 & 65.9 & 86.4 & 89.1 & 38.0 & 98.5 & 73.1 \\
        CALE (MBERT) & 78.7 & 66.6 & 83.5 & 87.2 & 41.9 & 98.9 & 67.0 \\
        CALE (XL-LEX) & 79.2 & 67.2 & 84.4 & 87.2 & 43.8 & 98.9 & 69.4 \\
        \arrayrulecolor{black}\bottomrule
    \end{tabular}
    \caption{Accuracies on test pairs from SemCor (Balanced Accuracies are indicated with $\dagger$). The higher the better. ``All'' refers to all pairs, ``SL'' (resp. ``DL'') to pairs of occurrences of the same lemma (resp. of different lemmas) and ``SC'' (resp. ``DC'') to pairs of occurrences referring to the same concept (resp. to different concepts).}
    \label{tab:accuracies_full}
\end{table*}

\begin{table*}[h]
    \centering
    \resizebox{\textwidth}{!}{
    \begin{tabular}{lcccccccccccc}
        \toprule
         & \multicolumn{2}{c}{\textbf{EN}} & \multicolumn{2}{c}{\textbf{DE}} & \multicolumn{2}{c}{\textbf{ES}} & \multicolumn{2}{c}{\textbf{IT}} & \multicolumn{2}{c}{\textbf{LA}} & \multicolumn{2}{c}{\textbf{SV}} \\
         \cmidrule(lr){2-3}
         \cmidrule(lr){4-5}
         \cmidrule(lr){6-7}
         \cmidrule(lr){8-9}
         \cmidrule(lr){10-11}
         \cmidrule(lr){12-13}
         & APD & PRT & APD & PRT & APD & PRT & APD & PRT & APD & PRT & APD & PRT \\
        \midrule\arrayrulecolor{white}
        XLM-R (first) & .221 & .313 & .143 & .307 & .078 & .308 & .180 & .327 & .004 & .196 & .098 & .031 \\
        XLM-R (mid) & .568 & .500 & .543 & .756 & .495 & .510 & .316 & .538 & \textbf{.237} & .481 & .575 & .466 \\
        XLM-R (last) & .564 & .527 & .342 & .404 & .526 & .562 & .270 & .605 & .001 & .521 & .518 & .307 \\
        XL-LEXEME & \textbf{.786} & \textbf{.671} & .798 & .803 & .633 & .610 & .512 & .637 & .142 & \textbf{.549} & .854 & .739 \\
        \midrule
        \midrule
        CALE (XLM-R) & .705 & .533 & .777 & \textbf{.823} & .655 & .543 & \textbf{.755} & \textbf{.753} & .108 & .461 & .833 & .720 \\
        CALE (MBERT) & .655 & .527 & - & - & - & - & - & - & - & - & - & - \\
        CALE (XL-LEX) & .730 & .570 & \textbf{.826} & .799 & \textbf{.685} & \textbf{.639} & .543 & .676 & .195 & .485 & \textbf{.859} & \textbf{.766} \\
        \arrayrulecolor{black}\bottomrule
    \end{tabular}
    }
    \caption{Spearman correlations between gold change scores and predicted change scores for LSCD on the DWUG datasets, for APD and PRT measures.}
    \label{tab:dwugs-complete}
\end{table*}

\section{Hardware and Code}
We conducted all experiments with Nvidia A30 GPU card with 24GB memory and Intel Xeon
Gold 5320 CPU. The main libraries used include Pytorch 2.5.1, HuggingFace transformers 4.48.1, datasets 3.2.0 and sentence-transformers 3.3.1. Total training time for CALE models ranges from 16-20 hours including hyper-parameter search. Evaluation time ranges approximately from 1-2 hours.

\section{Scientific Artifacts}

We used WordNet and SemCor, both properties of Princeton University.
Licence can be found at \url{https://wordnet.princeton.edu/license-and-commercial-use}.

For DWUG datasets:
\begin{itemize}
    \item Italian and Latin datasets are licenced under CC-BY 4.0 (\url{https://creativecommons.org/licenses/by/4.0/}).
    \item English, German, Spanish, Swedish datasets are licenced under CC-BY-ND 4.0 (\url{ https://creativecommons.org/licenses/by-nd/4.0/}).
\end{itemize}

The CoSimLex dataset, distributed at \url{https://huggingface.co/datasets/cjvt/cosimlex}, is published under the GNU GPL 3.0 licence \url{https://www.gnu.org/licenses/gpl-3.0.html}.

The SPCD dataset was derivated from the material of SemCor and WordNet and is intended for research use.

\end{document}